\documentclass[numsec,webpdf,modern,large,  hyphens]{oup-authoring-template}

\graphicspath{{Fig/}}

\theoremstyle{thmstyleone}%
\theoremstyle{thmstyletwo}%
\theoremstyle{thmstylethree}%

\usepackage{xcolor}
\usepackage[autolanguage]{numprint}
\usepackage{booktabs}
\usepackage{hyperref}
\hypersetup{colorlinks,allcolors=black}
\usepackage{pdfpages}

\begin{document}

\journaltitle{Journal Title Here}
\DOI{DOI added during production}
\copyrightyear{YEAR}
\pubyear{YEAR}
\vol{XX}
\issue{x}
\access{Published: Date added during production}
\appnotes{Paper}

\firstpage{1}


\title[Linking Workflow Information across Papers and Code]{Supporting Workflow Reproducibility by Linking Bioinformatics Tools across Papers and Executable Code}

\author[1,$\ast$]{Clémence Sebe\ORCID{0000-0003-1988-1875}}
\author[2]{Olivier Ferret\ORCID{0000-0003-0755-2361}}
\author[1]{Aurélie Névéol\ORCID{0000-0002-1846-9144}}
\author[3]{Mahdi Esmailoghli\ORCID{0009-0009-2148-6402}}
\author[3]{Ulf Leser\ORCID{0000-0003-2166-9582}}
\author[1,4,$\ast$]{Sarah Cohen-Boulakia\ORCID{0000-0002-7439-1441}}

\address[1]{\orgdiv{Université Paris-Saclay, CNRS}, \orgname{Laboratoire Interdisciplinaire des Sciences du Numérique}, \orgaddress{ \postcode{91400}, \state{Orsay}, \country{France}}} 
\address[2]{\orgdiv{Université Paris-Saclay}, \orgname{CEA, List}, \orgaddress{\postcode{F-91120}, \state{Palaiseau}, \country{France}}} 
\address[3]{\orgdiv{Department of Computer Science}, \orgname{Humboldt-Universität zu Berlin}, \orgaddress{\postcode{10099}, \state{Berlin}, \country{Germany}}} 
\address[4]{\orgdiv{Institut Universitaire de France (IUF)}}

\corresp[$\ast$]{Corresponding author. \href{clemence.sebe@universite-paris-saclay.fr}{clemence.sebe@universite-paris-saclay.fr}, \href{sarah.cohen-boulakia@universite-paris-saclay.fr}{sarah.cohen-boulakia@universite-paris-saclay.fr}}

\received{Date}{0}{Year}
\revised{Date}{0}{Year}
\accepted{Date}{0}{Year}

\editor{Associate Editor: Name}

\abstract{
\textbf{Motivation:} The rapid growth of biological data has intensified the need for transparent, reproducible, and well-documented computational workflows. The ability to clearly connect the steps of a workflow in the code with their description in a paper would improve workflow comprehension, support reproducibility, and facilitate reuse. This task requires the linking of bioinformatics tools in workflow code with their mentions in a published workflow description. \\
\textbf{Results:} We present CoPaLink, an automated approach that integrates three components: named entity recognition (NER) for identifying tool mentions in scientific text, NER for tool mentions in workflow code, and entity resolution based on word embedding similarity. We propose approaches for all three steps, achieving a high individual F1-measure (77 - 90) and a joint accuracy of 66 when evaluated on Nextflow workflows using Sentence-BERT. CoPaLink leverages corpora of scientific articles and workflow executable code with curated tool annotations to bridge the gap between narrative descriptions and workflow implementations. \\
\textbf{Availability:} The code is available at \url{https://gitlab.liris.cnrs.fr/sharefair/copalink-experiments} and \url{https://gitlab.liris.cnrs.fr/sharefair/copalink}. The corpora are also available: CPL-Article (\url{https://doi.org/10.5281/zenodo.20746904}), CPL-Code (\url{https://doi.org/10.5281/zenodo.20746970}) and CPL-Gold-Entity-Resolution (\url{https://doi.org/10.5281/zenodo.20746994}).\\
\textbf{Contact:} \href{clemence.sebe@universite-paris-saclay.fr}{clemence.sebe@universite-paris-saclay.fr}, \href{sarah.cohen-boulakia@universite-paris-saclay.fr}{sarah.cohen-boulakia@universite-paris-saclay.fr}\\
}


\maketitle

\section{Introduction}
With the rapid development of biomedical research technologies, the volume and diversity of generated biological data have grown dramatically. Processing this data has become essential to support the transition from raw data to meaningful knowledge and thus to enable discoveries. Ensuring reproducibility and thorough documentation of analysis pipelines is crucial to guarantee full traceability of data processing. Scientific workflow systems have emerged to address this need, enabling the design and execution of reproducible, transparent, and well-documented workflows \citep{cohenboulakia}.

Snakemake \citep{snakemake} and Nextflow \citep{nextflow} are some of the most widely adopted workflow management systems in the bioinformatics community. These systems are \textit{code-based}, i.e., workflows are defined as program code that orchestrates calls to bioinformatics tools, manages distributed and parallel execution, and provides specific configurations. As a result, workflows are commonly represented and shared as executable code, hosted on general-purpose code platforms such as GitHub or GitLab, or in workflow-specific repositories such as WorkflowHub \citep{gustafsson2025workflowhub} or nf-core \citep{ewels_nf-core_2020, langer_empowering_2025}.

However, to properly reuse workflows, it is necessary to understand recommended usage scenarios, implicit constraints, and limitations. Furthermore, other researchers usually want to assess the scientific results obtained through a workflow as a basis for their own choices. This information is usually not available in the workflow itself but rather in scientific publications that describe workflows in natural language text, often providing a link to the code repository.
For instance, as of January 2026, PubMed Central lists 3,858 papers mentioning Nextflow, including 2,923 with a linked GitHub repository.

Because workflows are described very differently in publications and in executable code, directly linking workflow descriptions in the literature with their executable code is highly challenging. Sources of deviations are manyfold, ranging from different naming of tools (e.g., \texttt{CircularMapper} versus \texttt{realignsamfile}), omitted steps in the paper (e.g., necessary format conversions or filtering operations), undescribed handling of extreme cases, or undocumented test routines to changes in the repository due to further development of a workflow after its publication. 
An important step in any attempt to match the workflows in papers to their code in a repository is the linking of the individual steps or tools within a workflow.

The software tools invoked within a workflow play a central role in defining its computational behavior. Ensuring that the tools mentioned in a paper correspond to those actually executed in the code is therefore essential for fully understanding a workflow and how it could be reused.

Linking tool mentions across executable code and textual descriptions can be formulated as an entity resolution (ER) task \citep{getoor_entity_2012,christophides_overview_2020}. ER usually aims to select and match occurrences of the same entity across different databases. In this context, entities are generally defined by structured attribute-based representations. In our work, we extend ER to the case where the entities to match come from two different media, text and source code, and appear as mentions in context. A related but distinct area of research is entity linking (EL), which associates entities mentioned in text with corresponding entries in a knowledge base (KB). When domain-specific KBs are available, EL can provide representations to facilitate entity resolution across heterogeneous sources.

In this work, we introduce CoPaLink, an entity resolution approach that automatically links tool mentions across the two complementary representations of a workflow: executable code and the associated publication. 
Our setting differs from conventional ER tasks because entity mentions are observed in two complementary representations of the same workflow. We investigate two complementary strategies: (i) direct entity resolution across the two representations, and (ii) indirect resolution through a shared external knowledge base when available. We refer to this setting as \textit{intermodal entity resolution}.
Furthermore, as the two sources are different genres of token sequences, the task can also be viewed as cross-modal alignment. Recent progress has been made in cross-modal alignment, especially in aligning textual and visual data (e.g., \cite{yarom2023you}). However, as far as we know, no existing work addresses alignment between text and workflow code at the level of individual workflow steps.

CoPaLink addresses two steps: NER and entity disambiguation in a pipeline approach. In the NER step, we introduce two novel corpora specifically focused on bioinformatics tools, building upon the BioToFlow corpus \citep{sebe_ida_2025}. Existing resources for software mention detection, such as Softcite \citep{du_softcite_2021}, primarily target generic biomedical software and do not explicitly model workflow-level bioinformatics tools. Prior methodological approaches range from rule-based systems \citep{duck_bionerds_2013} to supervised models \citep{wei_software_names} and workflow reconstruction frameworks \citep{halioui_bio_extract,BioWorkflow}. 
However, these efforts focus on generic software mentions or on full workflow reconstruction, rather than on structured alignment between the tools cited in articles and those used in executable code. In contrast, CoPaLink targets domain-specific bioinformatics tools and combines supervised NER with entity disambiguation based on word embedding similarity to systematically link literature and software repositories.

CoPaLink has been evaluated against a gold standard of 490 tools across 39 Nextflow workflows. Accuracy ranges from 56 to 84, with an average of 66, indicating its ability to reconcile tool mentions across papers and executable code.

\section{Methods}
\begin{figure*}[!ht]
\centering
    \includegraphics[width=.70\textwidth]{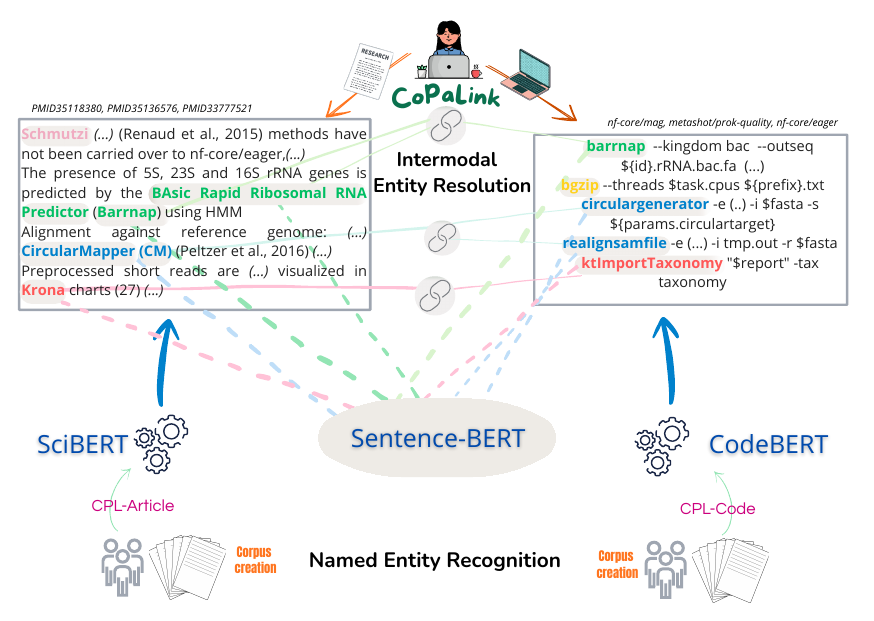}
    \caption{Architecture of CoPaLink. Bioinformatics tools are extracted from two sources (articles and code) using NER methods and then aligned using KBs. 
    }
    \label{fig.graph_ab}
\end{figure*}

This section introduces the components of CoPaLink shown in Figure \ref{fig.graph_ab}. Bioinformatics tool mentions are identified in text (left) and code (right) using supervised named entity recognition that leverages curated annotated corpora and language models. Mentions are linked with the support of domain KBs.

\subsection{Corpus creation}

We describe the process followed to create the corpora and the gold standards for each task. Both corpora, namely CPL-Article and CPL-Code, consist of scientific documents and workflow code manually annotated with mentions of bioinformatics tools.

\subsubsection{Gold standard for named entity recognition}
\paragraph{Source Material Selection}
Publications were initially obtained from the BioToFlow corpus, which contains 26 annotated papers describing Nextflow workflows~\citep{sebe_ida_2025}. More precisely, BioToFlow includes 15 papers, each with a link to a GitHub repository under an open license, and 11 papers with no exploitable code. This corpus was expanded by 24 papers and their associated code. To this end, we searched for workflows mentioning Nextflow in the title or abstract and associated with publicly available code, using the PubMed query \textit{nextflow[tiab] AND github[All Fields]} and the Python library Entrezpy \citep{Entrezpy}. Only articles published under a license permitting redistribution and reuse met our selection criteria. For the final corpus, CPL-Article, 50 papers were randomly selected, including 39 with associated code.

For the papers, \textit{Materials-and-methods}-type sections were identified and automatically extracted from each selected paper using an algorithm that matches section headings against predefined patterns (e.g., \textit{Methods}, \textit{Workflow}, \textit{Implementation}).
As for code, the Nextflow workflow code associated with each paper was retrieved from GitHub. We focused specifically on Nextflow \textit{processes}, as these are the units most likely to contain tool mentions. Process extraction was performed automatically using BioFlow-Insight~\citep{marchment_bioflow}.

\paragraph{Annotation Procedure}
The annotation procedure followed the state-of-the-art methodology described in~\cite{fort2016annotationBook} (Chapters 1--2) and previously implemented in~\cite{sebe_ida_2025} for the BioToFlow corpus. Both corpora were manually annotated according to detailed annotation guidelines specifying how to identify and categorize bioinformatics tool mentions in articles and executable code. A total of nine annotators contributed to the annotation process. Inter-annotator agreement (IAA)~\citep{artsteinCL2008} was measured throughout, and regular discussions were held to resolve ambiguities and refine the guidelines. All annotations were performed using BRAT (Brat Rapid Annotation Tool~\citep{stenetorp2012brat}). Both corpora are released in BRAT format.

\subsubsection{Gold standard for entity resolution}
Gold standard links between tool mentions in text and code are curated in the CPL-Gold-Entity-Resolution corpus.
For each workflow described in CPL-Article, we extracted a single mention of every tool extracted in either the text or the corresponding executable code. We then manually annotated correspondences between tools across the two modalities by constructing lists of tool pairs. For each tool identified in an article, we annotated its corresponding match(es) in the executable code, and vice versa. When no correspondence existed, the tool was explicitly annotated as unlinked. Table~\ref{tab.cpl-gold} represents the CPL-Gold-Entity-Resolution related to the example in Figure~\ref{fig.graph_ab}.

\begin{table}[!ht]
    \caption{\label{tab.cpl-gold} Example annotation for CPL-Gold-Entity-Resolution}.
    \centering
    \resizebox{\columnwidth}{!}{
    \begin{tabular}{ll}
    \toprule
    Article & Executable code \\
    \midrule 
    Schmutzi & \_ \\
    BAsic Rapid Ribosomal RNA Predictor & barrnap \\
    Barrnap & barrnap \\
    CircularMapper & circulargenerator \\
    CircularMapper & realignsamfile \\ 
    CM & circulargenerator \\
    CM & realignsamfile \\ 
    \_ & bgzip \\ 
    Krona & ktImportTaxonomy \\
    \bottomrule
    \end{tabular}}
\end{table}

\subsection{Named entity recognition}
To automatically identify bioinformatics tool names, we experiment with multiple NER approaches: first an annotation-free approach based on a KB (Section~\ref{sec_dico_approach}); then, two training-free approaches (Sections~\ref{sec_propa_approach} and \ref{sec_decoder_approach}); finally, an encoder-based supervised model (Section~\ref{sec_encoder_approach}) and the early fusion of this model and the KB of the first approach (Section~\ref{sec_encoder_approach_voc}).

\subsubsection{KB-based approach}\label{sec_dico_approach}
We used four knowledge bases (Biotools, Bioconda, Biocontainers, and Bioweb) to extract bioinformatics tool names and, when available, their commands, building a comprehensive dictionary. This dictionary was mapped onto the text (CPL-Article) and the processes (CPL-Code).

\subsubsection{Propagation approach (rote classifier)} \label{sec_propa_approach}

To determine the impact of memorization on entity extraction, we evaluated a baseline model that projected entities from the training to the test corpus using \citep{grouin_controlled_2016}.

\subsubsection{Decoder-based approach}\label{sec_decoder_approach}
\begin{figure}[!ht]
\centering
    \includegraphics[width=\columnwidth]{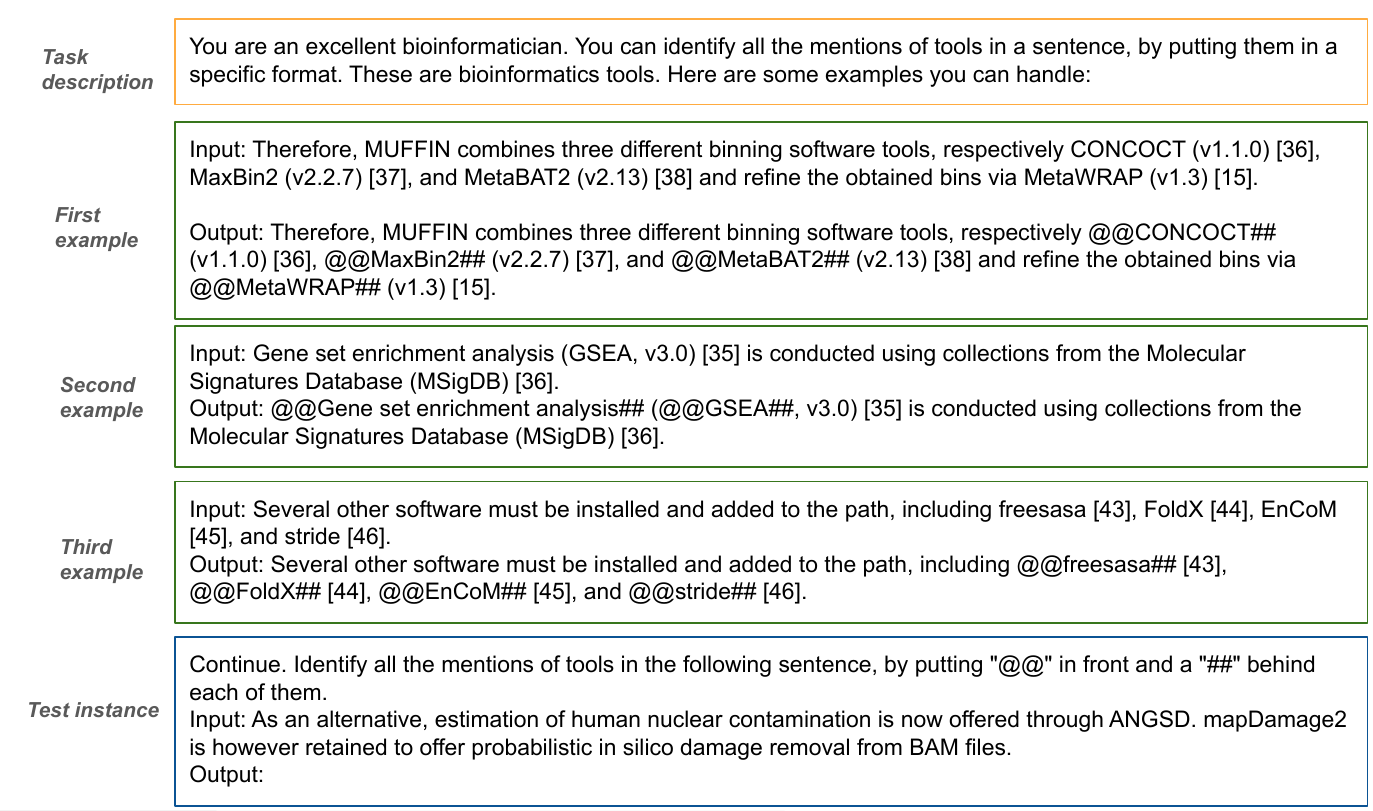}
    \caption{Example of a prompt for identifying tools in articles.}
    \label{fig.prompt}
\end{figure}

We used auto-regressive language models for few-shot NER. The methodology follows \cite{naguib_few-shot_2024}\footnote{\url{https://github.com/marconaguib/autoregressive_ner}} and experiments with different prompt formats and numbers of examples provided to the model.
Figure \ref{fig.prompt} presents a sample prompt used in our experiments. Prompts include three parts: (1) a description of the entity recognition task, which may include a definition of the bioinformatics tool entity (2) 5 to 10 examples; for clarity, only three examples are shown in Figure \ref{fig.prompt}. In the CPL-Article corpus, examples are complete sentences, whereas in the CPL-Code corpus, which does not contain natural language sentences, examples consist of four-line code snippets.
(3) the test instance to be analyzed by the model. For both corpora, several versions of Llama and Qwen models were evaluated.

\subsubsection{Encoder-based approach}\label{sec_encoder_approach}
Encoder-based approaches were evaluated using Transformer-based encoders (e.g., SciBert \citep{beltagy-etal-2019-scibert}, ModernBERT \citep{warner-acl-25}) coupled with a BiLSTM-CRF architecture \citep{huang_bidirectional_2015}  (see Figure 1 in Supplementary Data). The BiLSTM component captures contextual dependencies in both directions of a sequence, while the CRF layer models label dependencies in order to ensure coherent entity predictions.  We implemented this approach using the Python library NLStruct \citep{wajsburt_extraction_2021}\footnote{\url{https://github.com/percevalw/NLStruct}}, which supports multiclass classification and nested entity detection, and accepts brat-formatted input files.

\subsubsection{Encoder with vocabulary expansion} \label{sec_encoder_approach_voc} 
Finally, we tested the early fusion of the KB-based and encoder-based approaches, which can be considered complementary. 
The list of tools described in section~\ref{sec_dico_approach}
were used to expand the vocabulary of the language model of the encoder for both CPL-Article and CPL-Code.
The embeddings of newly introduced words were initialized as the average of the embeddings of their constituent existing words, like in \cite{gee-etal-2022-fast}.
These new embeddings were then adapted to the target context during encoder fine-tuning.


\subsection{Intermodal entity resolution}
To establish links between bioinformatics tool names across the two sources
(text and executable code), we 
compare methods ranging from string-to-string comparison (Section \ref{sec_sts_comparison}), through the incorporation of KBs (Section \ref{sec_kb}), to context-aware neural models (Sections \ref{sec_embedding} and \ref{sec_decoder}).

\subsubsection{String-to-string comparison} \label{sec_sts_comparison}
Our first strategy requires no external knowledge and simply performs a string-to-string comparison of tool names. A link is established only when an exact match is found between a tool mention in text and a tool mention in code. 
Two variants were considered: (i) Levenshtein distance, where two strings have a distance of $n$ if they differ by $n$ characters, and (ii) prefix–suffix matching, where two strings sharing the same prefix or suffix are considered identical.

\subsubsection{Knowledge bases as a bridge}\label{sec_kb}
KBs described in Section \ref{sec_dico_approach} are used as intermediaries: tool mentions are first mapped to KB entities, which 
serve as pivots to establish links between 
text and code. 
In KBs, bioinformatics tools are described using various attributes, including tool names and binary names. These alternative names
can be leveraged to identify different ways to refer to the same tool. 
Table \ref{ex_link_kb} illustrates examples of links between tool names mentioned in scientific articles and those appearing in executable code, as linked through the Bioconda KB.
In this setting, links are established using exact string matching over the alternative names associated with each KB entity.
Among the four KBs considered, only Bioconda, Biocontainers, and Bioweb provide multiple alternative names for the same tool, thereby enabling this type of systematic linking.

\begin{table*}[!ht]
    \centering
    \caption{\label{ex_link_kb} Example of links between tool names obtained from a KB.}
    \begin{tabular}{lll}
    \toprule
    Article & KB - Bioconda & Executable code \\
    \midrule
    circularmapper &  circularmapper, circulargenerator, realignsamfile  & circulargenerator \\
    rsem &  rsem, rsem-prepare-reference, rsem-bam2wig, \textit{etc.}  & rsem-prepare-reference \\
    metabat2 & metabat2, jgi\_summarize\_bam\_contig\_depths, contigOverlaps & jgi\_summarize\_bam\_contig\_depths \\ 
    \bottomrule
    \end{tabular}
\end{table*}

\subsubsection{Word embedding similarity} \label{sec_embedding}

We evaluated several alternative intermodal ER strategies based on semantic representations, using embeddings to capture lexical variation and contextual information beyond string matching. We experimented with two pre-trained language models, ModernBERT and Sentence-BERT \citep{reimers-2019-sentence-bert}\footnote{\url{https://huggingface.co/sentence-transformers/all-MiniLM-L6-v2}}. For ModernBERT, domain-specific vocabulary was injected. No model was fine-tuned, as training or fine-tuning an embedding-based ER model requires substantial annotated data to achieve competitive performance \citep{mehrafarin-etal-2022-importance}. Nevertheless, this approach allows us to leverage semantic representations learned from large corpora while avoiding overfitting on a limited annotated dataset. Tool representations were obtained either by encoding tool names alone or by encoding the textual context containing the tool mention, from which the corresponding tool embeddings were extracted. Then, candidate entities were compared according to cosine similarity in the embedding space. These approaches were evaluated both with and without KB-based candidate generation.

\subsubsection{Decoder-based approach} \label{sec_decoder}

Finally, we explored decoder-based approaches as an alternative strategy for ER under two configurations. 
In the first, the model was provided with all bioinformatics tool mentions extracted from both sources and was asked to predict the corresponding entity links, without any additional contextual information. In the second configuration, contextual information was explicitly included in the prompt. For each tool extracted from articles, a single prompt was constructed by aggregating all sentences in which the tool appears, followed by the list of all processes identified in the executable code. The model was then instructed to identify the correct matching entity.


\section{Corpus Statistics}

\subsection{CoPaLink corpora}
\textbf{CPL-Article} corpus is composed of 50 papers describing Nextflow bioinformatics workflows. IAA score reached 80\%, indicating a good understanding of the annotation guidelines. In preparation for NER experiments, the CPL-Article corpus was first split into a fixed test set with 25\% of the workflows (14) and a development set containing the remaining 75\%. To reduce dependence on a particular train/validation split, we generated five random splits from the development set. For each set, 75\% of the development data were used for training and 25\% for validation.

For consistency, the same data partitioning ratios were used for the \textbf{CPL-Code} corpus, which contains \numprint{1368} processes. The test set consists of processes extracted from the same 14 workflows used for CPL-Article evaluation, yielding a total of 377 processes. This design ensures that both corpora are evaluated on the same workflows. 
IAA on the CPL-Code corpus was 80\%, indicating strong agreement in the identification of bioinformatics tools. \\

\begin{table*}[!ht]
\caption{\label{stat_corpus}CPL-Code and CPL-Article statistics}.
\centering
\begin{tabular}{lcccccc}
\toprule
Corpus & Part & No. papers or code processes & No. words & No. words annotated & No. tools & No. unique tools \\ \midrule
CPL-Article & All & \numprint{50} & \numprint{76002} & \numprint{1523} & \numprint{1455} & \numprint{425} \\
    & Train/Val & \numprint{36} & \numprint{46701} & \numprint{757} & \numprint{693}  & \numprint{261}  \\
    & Test  & \numprint{14}  & \numprint{29301}  & \numprint{766} & \numprint{762}  & \numprint{216}  \\
\midrule
CPL-Code & All & \numprint{1368} & \numprint{137337} & \numprint{3401}  & \numprint{3398} & \numprint{537} \\
    & Train/Val & \numprint{991} & \numprint{94629}  & \numprint{2281} & \numprint{2278}  & \numprint{386}  \\
    & Test  & \numprint{377} & \numprint{42708}  & \numprint{1120} & \numprint{1120}  & \numprint{225}   \\
\bottomrule
\end{tabular}
\end{table*}

\noindent

Table \ref{stat_corpus} summarizes the statistics of the two corpora, including the number of annotated papers, code processes, and entities, and provides separate details for the training/validation and test sets. 
Tool overlap between development and test sets is moderate: CPL-Article contains 52 overlapping tools (\numprint{24.1}\% of unique test set tools, \numprint{26.7}\% of all test set tool mentions) and CPL-Code 74 (\numprint{32.9}\% unique, \numprint{63.5}\% all mentions). This indicates that most test set tools are unseen during training, supporting the use of these corpora to evaluate model generalization. In addition, we conduct a propagation experiment to assess whether this overlap affects performance (Section \ref{sec.res.propa}).

\subsection{Gold standard for entity resolution} 
Among the 50 workflows represented in CPL-Article, 39 also contain annotated tools in their corresponding executable code. 
These were selected to create the CPL-Gold-Entity-Resolution corpus, extract both bioinformatics tools named in papers and executable code, and then create the list of tool pairs. 
To assess annotation reliability, 12 workflows (including both articles and code) were independently annotated by two bioinformaticians. The annotation time varied across workflows, averaging approximately 30 minutes per workflow. IAA reached 92\%, indicating a high level of agreement. 
In total, CPL-Gold-Entity-Resolution comprises 39 workflows and 490 cross-modal tool links. Additionally, 125 tools are annotated as appearing exclusively in articles, and 195 exclusively in source code, reflecting differences in reporting practices between publications and code.

\section{Experimental Results}

\subsection{Metrics}
Task performance was evaluated using precision (P), recall (R), and F1-measure (F1) as defined below. 
\[ P  = \frac{TP}{TP + FP}, \quad R  = \frac{TP}{TP + FN}, \quad F1 = 2 \cdot \frac{P \cdot R}{P + R}\]

For NER tasks, TP (true positives) corresponds to correct tool predictions, FN (false negatives) to missed tool instances, and FP (false positives) to incorrect tool predictions. 
We used the BRAT-Eval implementation  (\cite{verspoor_annotating_2013}, v0.0.2). 
For the intermodal 	ER task, TP corresponds to correct link predictions, either linking two corresponding tool mentions or correctly leaving a tool mention unlinked when no match exists across modalities. FN corresponds to ground-truth links or unlinked cases missed by the approach, whereas FP corresponds to incorrect link predictions.

\subsection{Named entity recognition}

\subsubsection{KB-based approach} \label{sec_res_dico_approach}

This method provides a baseline as it does not involve any training model. Overall, performance varies substantially across KBs (a detailed Table is provided in Supplementary Data B). For CPL-Article, all approaches achieve relatively low F1-scores, with Bioconda performing best (\numprint{45.7}). These results highlight the difficulty of KB-based extraction in publications, where tool mentions are more diverse. 
In contrast, Biocontainers achieve the highest F1-score (\numprint{47.5}) on CPL-Code. Biocontainers exhibits high precision but limited recall, which can be explained by the relatively small size of the KB (approximately 900 tools\footnote{\url{https://github.com/BioContainers/containers}}).
Merging all KBs substantially increases recall in both subsets, particularly for CPL-Code. However, this gain comes at the cost of a sharp decrease in precision, resulting in no overall improvement in F1-score. 
In both cases, combining heterogeneous KBs offers increased coverage, though ambiguous entries may introduce some noise, highlighting the importance of KB quality in addition to size.

\subsubsection{Propagation approach (rote classifier)}\label{sec.res.propa}

\begin{table}[!ht]
\caption{\label{table.ner_propa}Performance obtained using a propagation approach.}
\centering
\begin{tabular}{lccc}
\toprule
                 & Precision & Recall & F1   \\ \midrule
CPL-Article  &  \numprint{91.0} & \numprint{26.7} & \numprint{41.2} \\
CPL-Code         & \numprint{27.9}  & \numprint{63.5} & \numprint{38.7} \\ \bottomrule
\end{tabular}
\end{table}

Using a simple annotation propagation approach from the training to the test sets, we obtain F1-scores of \numprint{41.2} for CPL-Article and \numprint{38.7} for CPL-Code (Table \ref{table.ner_propa}). CPL-Article exhibits very high precision (\numprint{91.0}) but low recall (\numprint{26.7}), indicating that propagated annotations are usually correct but sparse. In contrast, CPL-Code shows higher recall (\numprint{63.5}) but lower precision (\numprint{27.9}), suggesting that tool mentions in code are used for other purposes than tool use (e.g., in process or variable names). Despite the observed overlap between training and test sets, the relatively modest F1-scores indicate that simple memorization or annotation propagation is insufficient to explain the performance of more advanced methods on these corpora, supporting their suitability for evaluating generalization beyond previously seen tools.

\subsubsection{Decoder-based approach} \label{sec_res_decoder_approach}
\begin{table*}[!ht]
\caption{\label{tab.llm} Performance of bioinformatics tool name extraction in terms of precision, recall, and F-measure (F1) using the decoder-based approach.}
\centering
\begin{tabular}{llccc}
\toprule
Corpus & Model    & Precision & Recall & F1  \\  
\midrule
CPL-Article & Llama-3.1-8B-Instruct  \citep{grattafiori2024llama3herdmodels}              & \numprint{63.8} & \numprint{78.2} & \numprint{70.3} \\
& Qwen3-8B  \citep{qwen3technicalreport}                           & \numprint{65.6} & \numprint{73.7} & \numprint{69.4} \\ 
\midrule
CPL-Code & Llama-3.1-70B-Instruct \citep{grattafiori2024llama3herdmodels}              & \numprint{50.7} & \numprint{21.2} & \numprint{29.9} \\ 
& CodeLlama-13b-Instruct-hf  \citep{roziere2024codellamaopenfoundation}          & \numprint{40.5} & \numprint{13.7} & \numprint{20.5} \\
& Qwen3-8B  \citep{qwen3technicalreport}                            & \numprint{43.4} & \numprint{18.2} & \numprint{25.6} \\
& Qwen2.5-Coder-14B-Instruct  \citep{hui2024qwen2,qwen2}         & \numprint{63.5} & \numprint{20.3} & \numprint{30.7} \\
\bottomrule
\end{tabular}
\end{table*}

The results reported in Table \ref{tab.llm} show that decoder-based models achieve their best performance on the CPL-Article corpus but exhibit substantially weaker performance on CPL-Code. Table \ref{tab.llm} reports only the best-performing results for each model family; a more detailed Table is provided in Supplementary Data. On CPL-Article, Llama-3.1-8B-Instruct slightly outperforms Qwen3-8B. Performance drops on the CPL-Code corpus, where all evaluated models reach F1-scores below 35.
Among the code-oriented models, Qwen2.5-Coder-14B-Instruct achieves the highest F1-score, but overall gains remain limited.

These ﬁndings are consistent with previous work showing that decoder-based models are generally not well-suited for this task \citep{keraghel_recent_2024, naguib_few-shot_2024}.
As they are primarily optimized for next-token prediction rather than explicit NER, their ability to reliably identify and generalize bioinformatics tool names remains limited, particularly in code-centric contexts. However, we cannot exclude the possibility that the prompting strategy used for tool extraction from source code is not optimal, as code structure differs from that of natural sentences.

\subsubsection{Encoder-based approach} \label{sec_res_encoder_approach}
\begin{table}[!ht]
\caption{\label{encoder_res} Average performance metrics on five splits and five random seeds (with standard deviation) using NLStruct.}
\centering
\resizebox{\columnwidth}{!}{
\begin{tabular}{llccc}
\toprule
 Corpus & Model   & Precision   & Recall  &  F1 \\ 
 \midrule
CPL-Article & SciBERT  &            \numprint{84.2}{\scriptsize $\pm$\numprint{1.4}}  & \numprint{69.6}{\scriptsize $\pm$\numprint{1.9}} & \numprint{76.2}{\scriptsize $\pm$\numprint{0.9}} \\
& ModernBERT  &         \numprint{73.8}{\scriptsize $\pm$\numprint{2.1}}  & \numprint{69.9}{\scriptsize $\pm$\numprint{1.9}} & \numprint{71.8}{\scriptsize $\pm$\numprint{0.9}}  \\
& \textbf{SciBERT + injection} & \textbf{\numprint{87.9}{\scriptsize $\pm$\numprint{2.2}}}  & \textbf{\numprint{69.0}{\scriptsize $\pm$\numprint{3.5}}} & \textbf{\numprint{77.2}{\scriptsize $\pm$\numprint{1.8}}} \\
\midrule
CPL-Code & CodeBERT   & \numprint{91.1}{\scriptsize $\pm$\numprint{0.8}} & \numprint{88.0}{\scriptsize $\pm$\numprint{0.9}} & \numprint{89.5}{\scriptsize $\pm$\numprint{0.6}} \\ 
& SciBERT    & \numprint{88.4}{\scriptsize $\pm$\numprint{0.8}} & \numprint{85.9}{\scriptsize $\pm$\numprint{0.7}} & \numprint{87.1}{\scriptsize $\pm$\numprint{0.5}}  \\ 
& ModernBERT & \numprint{88.9}{\scriptsize $\pm$\numprint{1.1}} & \numprint{89.0}{\scriptsize $\pm$\numprint{0.8}} & \numprint{89.0}{\scriptsize $\pm$\numprint{0.4}} \\ 
& \textbf{CodeBERT + injection} & \textbf{\numprint{91.4}{\scriptsize $\pm$\numprint{0.8}}} & \textbf{\numprint{88.8}{\scriptsize $\pm$\numprint{1.2}}} & \textbf{\numprint{90.1}{\scriptsize $\pm$\numprint{0.6}}} \\
\bottomrule
\end{tabular}}
\end{table}

For each model, training was repeated using five train/validation splits and multiple random seeds to ensure robustness and reproducibility. The reported results correspond to the average performance of each model on the test set and are reported in Table \ref{encoder_res}. For the CPL-Article corpus, three models were tested. We first evaluated two pre-trained models: SciBERT, which is trained on scientific literature, and ModernBERT, a recent and optimized version of BERT. This initial comparison shows that SciBERT performs better than ModernBERT in extracting bioinformatics tools in articles. The Almost Stochastic Order (ASO) test \citep{dror_deep_2019} was applied at a \numprint{0.05} confidence level to assess the statistical significance among models. The results indicate that SciBERT is stochastically dominant over ModernBert ($\epsilon_{min}=0$). As for the third model considered, we extended SciBERT by incorporating domain-specific bioinformatics vocabulary and fine-tuning it for the target task.

For the CPL-Code corpus, we compared CodeBERT (pre-trained on code) \citep{feng2020codebert}, SciBERT (on literature), and ModernBERT (on both). Their F1-scores are very close, suggesting process code can be treated like natural language; CodeBERT performs slightly better. However, ASO testing shows no significant differences, especially between ModernBERT and CodeBERT. As with CPL-Article, we further extended CodeBERT with domain-specific bioinformatics vocabulary, yielding a slight but non-significant gain.

\subsection{Intermodal entity resolution}

We present the results of the tool linking 
using the 39 workflows of the CPL-Gold-Entity-Resolution corpus, comparing bioinformatics tool names in the articles with those present in the executable code at the time of publication.

\subsubsection{String-to-string comparison} \label{sec_res_el_sts}
We conducted different experiments using string-to-string comparison. A Table with all the results is available in Supplementary Data (C.1).
Our first approach, based on exact string matching, achieves an F1-score of \numprint{75.1}, highlighting the limitations of exact matching in the presence of naming variations. The other string-matching approaches yield slightly better performance than the initial approach. Because tool names are often short, increasing the Levenshtein distance threshold beyond three reduces performance, whereas thresholds of one or two yield the best results, with a maximum F1-score of \numprint{76.8}.
Prefix- and suffix-based matching strategies were also explored, but their performance remained comparable to the other string-matching approaches, reaching a maximum F1-score of \numprint{78.0}.

\subsubsection{Using knowledge bases as a bridge} \label{sec_res_el_kb}
The performance obtained using the different KBs, as well as their combinations through transitive grouping of tool names, is reported in Table \ref{table.el_table}. Only the best KB fusions are shown. Overall, the fusion of Bioconda and Bioweb achieves the best performance on CPL-Gold-Entity-Resolution, with the highest F1-score of \numprint{81.6}. Among individual KBs, Bioconda-exact performs best, reaching an F1-score of \numprint{80.3}, closely followed by Bioweb-exact (\numprint{80.1}).
The fusion of all KBs does not yield a significant improvement over Bioconda-exact alone, suggesting that most of the valuable information is already covered by Bioconda. However, combining Bioconda with Bioweb yields a slight gain, indicating complementary coverage between the two sources. 

\begin{table}[!ht]
\caption{\label{table.el_table} Performance using KB to do the linking in CPL-Gold-Entity-Resolution. Exact string matching used is expressed by -exact.}
\centering
\resizebox{\columnwidth}{!}{
\begin{tabular}{lccc}
\toprule
              & Precision & Recall & F1    \\ 
\midrule
Bioconda-exact      & \numprint{79.7}     & \numprint{81.0}  & \numprint{80.3} \\ 
Biocontainers-exact & \numprint{77.4}     & \numprint{78.3}  & \numprint{77.8} \\
Bioweb-exact        & \numprint{79.5}     & \numprint{80.7}  & \numprint{80.1} \\
\midrule
KB All fusion-exact     & \numprint{80.5}     & \numprint{81.4}  & \numprint{80.9} \\
\textbf{Bioconda-Bioweb-fusion-exact}     & \textbf{\numprint{81.2}}     & \textbf{\numprint{82.1}} & \textbf{\numprint{81.6}} \\ 
\bottomrule
\end{tabular}}
\end{table}

Finally, we combined our best KB-approach (Bioconda-Bioweb-fusion-exact) with string-to-string approaches. The results show that combining these approaches does not lead to any further improvement (see Supplementary data (C.2)).

\subsubsection{Word embedding similarity} \label{sec_res_el_emb}

The results obtained with embedding-based methods are presented in Table \ref{table.el_other_res}. Performance varies substantially across embedding models. While the ModernBERT-based approach remains below both the string-to-string baseline and the KB-based exact matching method, Sentence-BERT achieves the best overall performance. This approach outperforms the KB-based exact matching strategy, achieving higher F1 scores (\numprint{83.4} vs. \numprint{81.6}) while requiring less computation time (2 seconds vs. 20 seconds). This result indicates that Sentence-BERT produces embeddings that are better suited to our entity resolution setting.
The incorporation of contextual information significantly degrades performance across both embedding models. This suggests that the contextual information considered in our experiments introduces noise and does not improve disambiguation. For this task, the surface form of tool names appears to provide a stronger signal than the surrounding sentence context.

\begin{table*}[!ht]
\caption{\label{table.el_other_res} Performance obtained using word embedding similarity to do the linking in CPL-Gold-Entity-Resolution}.
\centering
\begin{tabular}{l@{\hskip 2em}ccc}
\toprule
     & Precision & Recall & F1 \\ \midrule
String-to-string & \numprint{74.6}     & \numprint{75.7}  & \numprint{75.1} \\
Bioconda-Bioweb-fusion-exact     & \numprint{81.2}    & \numprint{82.1} & \numprint{81.6} \\
\midrule
Embeddings similarity &  \numprint{60.9}         &  \numprint{70.2}      & \numprint{65.2} \\
Embeddings + Bioconda-Bioweb-fusion-exact   &   \numprint{66.7}        &  \numprint{72.6}      &  \numprint{69.5}  \\
\midrule
Embeddings+context   &  \numprint{28.7}         & \numprint{39.3}       & \numprint{33.1}   \\
Embeddings + context + Bioconda-Bioweb-fusion-exact    &    \numprint{4.8}        &  \numprint{59.6}      &  \numprint{9.0}   \\
\midrule
\textbf{Sentence-BERT}   &  \textbf{\numprint{83.2}}         & \textbf{\numprint{85.1}}       & \textbf{\numprint{84.1}}   \\
Sentence-BERT + context   &    \numprint{36.9}        &  \numprint{50.1}      &  \numprint{42.5}   \\ \bottomrule
\end{tabular}
\end{table*}

\subsubsection{Decoder-based approach} \label{sec_res_el_decoder}
We used the same models as in the NER experiments, excluding code-specific ones, since prompts combine natural language and code, making general-purpose models more suitable. As shown in Table~\ref{table.el_other_res_2}, decoder-based approaches, with or without context, perform worse than the string-to-string method overall. Without context, models achieve the lowest scores. Adding context consistently improves precision, recall, and F1, but results still remain below the string-to-string baseline.

\begin{table}[!ht]
\caption{\label{table.el_other_res_2} Performance obtained using decoder-based approaches to do the linking in CPL-Gold-Entity-Resolution}.
\centering
\resizebox{\columnwidth}{!}{
\begin{tabular}{l@{\hskip 3em}ccc}
\toprule
 Model    & Precision & Recall & F1 \\
 \midrule
\textbf{String-to-string} & \textbf{\numprint{74.6} }    & \textbf{\numprint{75.7}}  & \textbf{\numprint{75.1}} \\
Qwen3-8B   &  \numprint{54.2}         &  \numprint{34.8}      &  \numprint{42.4} \\
Llama-3.1-8B-Instruct        &   \numprint{49.1}        &  \numprint{40.2}      &  \numprint{44.2} \\
Qwen3-8B  + context  &  \numprint{59.7}  & \numprint{61.9} & \numprint{60.8} \\
Llama-3.1-8B-Instruct  + context  & \numprint{56.1} & \numprint{58.9}  & \numprint{57.4}  \\
\bottomrule
\end{tabular}}
\end{table}

\subsection{Full pipeline evaluation}
As a final evaluation, we assessed \textbf{CoPaLink} as a fully integrated, end-to-end system that leverages the best approaches at each step. Using the same 14 workflows previously employed in the NER experiments, the complete pipeline achieves an overall performance of \numprint{65.8}. This result demonstrates the feasibility of automatically linking tool mentions across executable workflow code and their associated scientific publications, while also highlighting the challenges of this task due to error propagation across pipeline components. \\

\section{Discussion}

This study introduces the first complete analysis pipeline designed to extract mentions of bioinformatics tools from both workflow code and papers and to identify links between them.
Our approach is applicable across various contexts and can benefit different user types. For researchers preparing a manuscript, it enables verification of the consistency between the workflow implementation and its description before submission. For readers of the paper under review or after publication, it facilitates workflow understanding and reuse by clearly identifying the tools involved and explicitly linking their occurrences across sources. Returning to the example illustrated in Figure \ref{fig.graph_ab}, our method can independently extract \texttt{CircularMapper} and \texttt{circulargenerator} from the two sources and subsequently reconnects them via Sentence-BERT.  
In the following, we discuss each step of our pipeline, place our contributions in the context of related work, and assess the approach's environmental impact in the Supplementary Data.

\subsection{Named entity recognition}
In our previous work on NER \citep{sebe_ida_2025}, we explored using existing corpora to extract workflow-related information from textual descriptions. We evaluated the integration of Softcite \citep{du_softcite_2021} and showed that incorporating this corpus did not improve extraction performance in our setting. This result suggests limited transferability between generic biomedical software corpora and domain-specific bioinformatics workflow entities. Therefore, we did not further evaluate additional generic software corpora relying on similar annotation principles and domain coverage, such as large-scale corpora derived from \cite{istrate_large_2022},  \cite{dannenfelser-neurips-23}, or \cite{SoMeSci}.

More broadly, prior workflow-oriented approaches focus on full workflow reconstruction \citep{halioui_bio_extract} or LLM-based (large language models) question answering \citep{BioWorkflow}. They address high-level reconstruction or querying, but not the span-level identification of tools needed for ER.

Methodologically, our experiments (Tables \ref{tab.llm}-\ref{encoder_res}) show that small supervised models augmented with domain vocabulary outperform KB-based and few-shot LLM approaches in both text and code. This aligns with earlier structured NER methods such as CRFs \citep{wei_software_names}. BiLSTM-CRF models extend CRFs by learning contextual representations from full text, improving generalization to rare, domain-specific tools. We go one step further by injecting external domain knowledge directly into a BiLSTM-CRF via vocabulary augmentation, an early-fusion strategy that enhances the detection of rare or unseen tools. A complementary late-fusion approach could retain only predictions matching a KB.

\subsection{Intermodal entity resolution}

Methods that incorporate more contextual information do not perform better than those without any context. This suggests that the characterization of the tool lies in lexical and semantic information conveyed by the tool name rather than the context around a tool mention. The optimal ER strategy in our experiments is the use of Sentence-BERT, a word-embedding approach (Table~\ref{table.el_other_res}).  
This can be explained by the nature of bioinformatics tool names, which are generally short and often highly similar, thereby limiting the contribution of contextual information for disambiguation. 
Indeed, embeddings convey semantic information about tool context on a broader scale than local context surrounding a single mention. Bioinformatics KBs, such as Biotools, rely on structured knowledge from EDAM \citep{edam} to build a broad yet more specialized representation of tools. However, this representation is heterogeneous across tools and is often limited by insufficient terminology coverage in a fast-paced field where new tools are released more frequently than KBs are updated.

\section{Conclusion and perspectives}
In summary, this work presents CoPaLink, an entity resolution approach that automatically aligns tool mentions across two complementary workflow representations: executable code and the associated scientific publication. While the relatively small size of our corpora may limit the generalizability of our findings, the results are strong and highlight the robustness of our methods as well as the contribution of domain-specific knowledge. In addition, because the corpus includes only Nextflow workflows, the results may reflect characteristics specific to this system. 

Regarding future work, several directions can be explored to further improve CoPaLink. For the NER component, we plan to investigate few-shot learning approaches, particularly by leveraging models capable of recognizing any type of entity, such as GLiNER \citep{zaratiana-etal-2024-gliner}. For the intermodal entity resolution step, increasing the amount of annotated training data and further fine-tuning the models represent promising directions, in line with recent work highlighting the role of entity matching across heterogeneous artifacts \citep{fuchs_whos_2026}. Additionally, we aim to strengthen the integration of domain knowledge by enhancing the embeddings of the injected vocabulary, notably by adapting them during NER model training, following approaches similar to those proposed in \cite{hong-avocado-21}. Finally, from a more general perspective, aligning the workflow description with its executable code is a promising direction for improving workflow reuse.

\section{Data availability}

Code is available: CoPaLink-Experiments (\url{https://gitlab.liris.cnrs.fr/sharefair/copalink-experiments}, \url{https://doi.org/10.5281/zenodo.20747813}) and CoPalink approach (\url{https://gitlab.liris.cnrs.fr/sharefair/copalink}, \url{https://doi.org/10.5281/zenodo.20747871}). Data are available: CPL-Article (\url{https://doi.org/10.5281/zenodo.20746904}, CPL-Code (\url{https://doi.org/10.5281/zenodo.20746970}) and CPL-Gold-Entity-Resolution  
(\url{https://doi.org/10.5281/zenodo.20746994}).

\section{Acknowledgments}

We thank N. Bossut, L. Buggenhoudt, A. Gaignard, S. Kaur, G. Marchment, H. Ménager, M. Schmit, F. Lemoine for their help with the annotation process, and M. Naguib for his help with the decoder-based experiments. We acknowledge E. Deveaud and the HPC Core Facility at Institut Pasteur for providing updates to Bioweb KB. This work was performed using HPC resources from GENCI-IDRIS (Grant 2025-AD011017259).

\section{Author Contributions}

A.N., O.F., and S.C.B. conceived the project.
C.S. implemented the methods, performed the experiments, and wrote the first draft of the manuscript. U.L. and M.E. provided advice for individual steps and analysis methods. All authors reviewed and edited the manuscript. Funding acquisition: S.C.B. and U.L.

\section{Funding}
This work has received support from the French government (Agence Nationale pour la Recherche) under the France 2030 program grant agreement ANR-22-PESN-0007 (ShareFAIR) and funding from the Deutsche Forschungs-gemeinschaft through the SFB 1404 FONDA (Project-ID 414984028).

\section{Conflicts of interest}
None declared


\includepdf[pages=-]{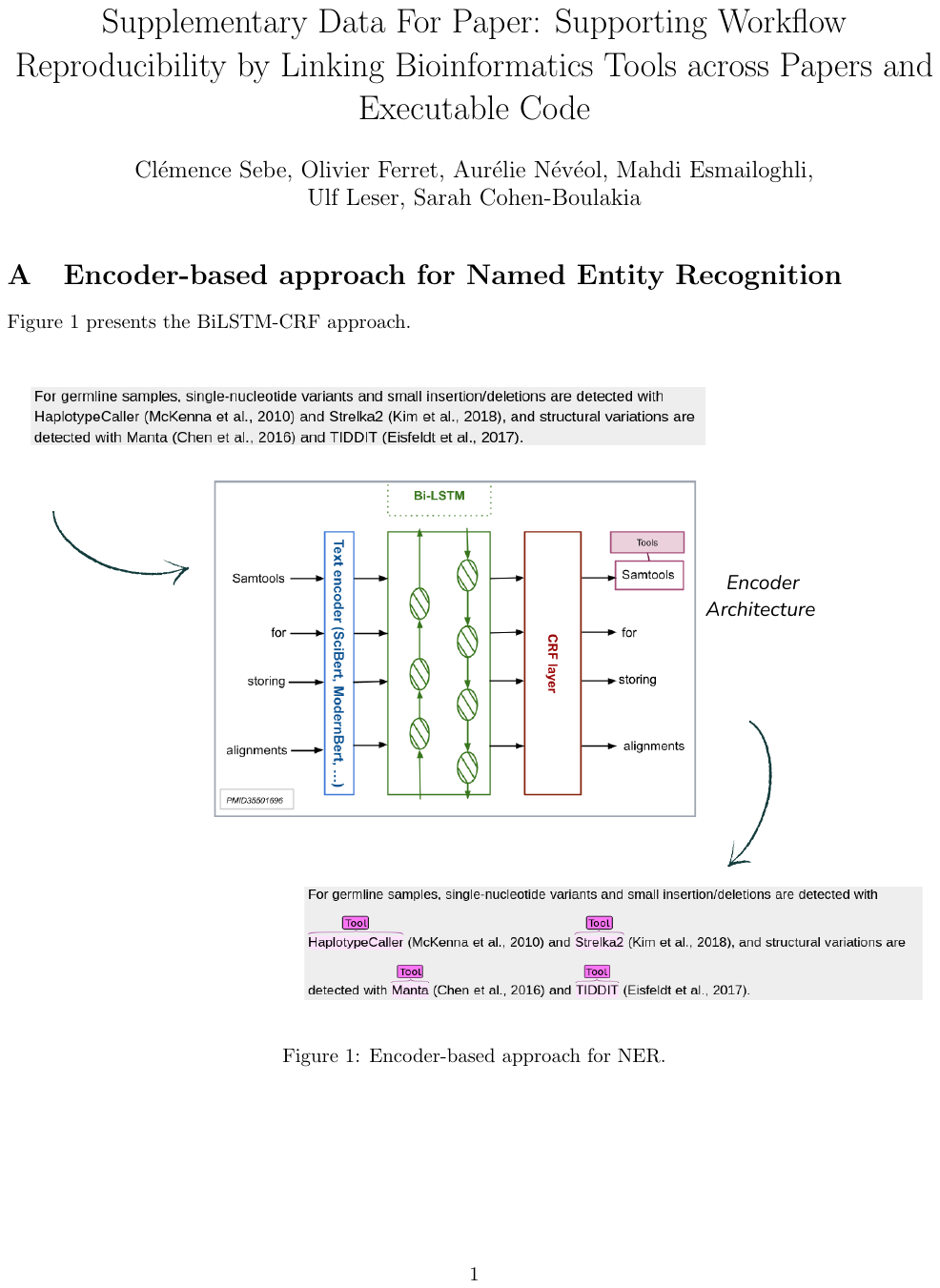}
\end{document}